\documentclass[conference]{IEEEtran}
\IEEEoverridecommandlockouts
\usepackage{times}
\usepackage{amsmath}
\usepackage{amssymb}
\usepackage{multicol,multirow}
\usepackage[tableposition=top]{caption}
\usepackage{cite}
\usepackage{booktabs}
\usepackage{bigstrut}
\usepackage{floatrow}
\usepackage{rotating}
\usepackage{adjustbox}
\usepackage{amsmath}

\def\BibTeX{{\rm B\kern-.05em{\sc i\kern-.025em b}\kern-.08em
    T\kern-.1667em\lower.7ex\hbox{E}\kern-.125emX}}
\begin{document}
\title{A Robust Algorithm for Contactless Fingerprint Enhancement and Matching}


\author{Mahrukh Siddiqui, Shahzaib Iqbal~\IEEEmembership{Member,~IEEE,}
  Bandar AlShammari,~\IEEEmembership{Member,~IEEE,}\\
  Bandar Alhaqbani,~\IEEEmembership{Member,~IEEE,}
        Tariq M. Khan,~\IEEEmembership{Member,~IEEE,}
        and~Imran Razzak,~\IEEEmembership{Member,~IEEE}

\thanks{Mahrukh Siddique is with the Visual Computing Technologies PVT, Islamabad, Pakistan (email: mahrukhsiddiqui34@gmail.com)}

\thanks{Shahzaib Iqbal is with the Department of Computing, Abasyn University Islamabad Campus(AUIC), Islamabad, Pakistan (email: shahzaib.iqbal91@gmail.com)}

\thanks{Bandar Alhaqbani and Bandar Alhaqbani are with TCC Research and Development Labs, Technology Control Company, Riyadh, Saudi Arabia (email: bmalshammari@tcc-ict.com; haqbanib@gmail.com)}

\thanks{Tariq M. Khan and Imran Razzak are with the School of Computer Science \& Engineering, UNSW, Sydney, Australia (e-mail: \{tariq.khan, imran.razzak\}@unsw.edu.au)}}  

\maketitle

\begin{abstract}
Compared to contact fingerprint images, contactless fingerprint images exhibit four distinct characteristics: (1) they contain less noise; (2) they have fewer discontinuities in ridge patterns; (3) the ridge-valley pattern is less distinct; and (4) they pose an interoperability problem, as they lack the elastic deformation caused by pressing the finger against the capture device. These properties present significant challenges for the enhancement of contactless fingerprint images. In this study, we propose a novel contactless fingerprint identification solution that enhances the accuracy of minutiae detection through improved frequency estimation and a new region-quality-based minutia extraction algorithm. In addition, we introduce an efficient and highly accurate minutiae-based encoding and matching algorithm. We validate the effectiveness of our approach through extensive experimental testing. Our method achieves a minimum Equal Error Rate (EER) of 2.84\% on the PolyU contactless fingerprint dataset, demonstrating its superior performance compared to existing state-of-the-art techniques. The proposed fingerprint identification method exhibits notable precision and resilience, proving to be an effective and feasible solution for contactless fingerprint-based identification systems.
\end{abstract}

\vspace{0.5\baselineskip}


\section{Introduction}
\label{sec:intro}

Contactless biometric technologies (such as identification and verification) have gained significant interest in commercial applications and are now a cutting-edge research area due to the advancement and popularity of sensing technologies\cite{michael2012contactless, michael2010innovative, oh2017gabor, zhou2014benchmark}. Standards and Technology National Institute (NIST) has declared that Next Generation Fingerprint Technologies will be developed.
One of the most significant components of this research is contactless fingerprint technology, which highlights the tremendous future potential of contactless fingerprint technologies.
The contactless fingerprint caption system can prevent numerous dangers, including picture contamination, time-consuming problems, non-linear distortion, and hygiene concerns compared to the contact-based fingerprint collection method. However, low ridge/valley contrast is often present in contactless fingerprint scans.
Many techniques have been put forth and made impressive strides in the last few decades to improve the ridge-valley contrast of fingerprint scans. Based on the filtering domain, the improvement techniques can be essentially divided into two groups: (1) spatial domain filtering \cite{khan2013fingerprint, khan2014fingerprint, khan2016spatial, khan2016stopping, khan2017efficient, khan2018coupling} and (2) Fourier domain filtering \cite{chikkerur2007fingerprint, sherlock1994fingerprint, watson1994comparison}. 
Contextual filters are the most commonly used spatial domain filtering techniques for improving fingerprint images. Contextual filters were first introduced for fingerprint improvement by Nickerson and O'Gorman \cite{nickerson1989approach}. The ridge frequency and ridge orientation regulate those filters \cite{sabir2020reducing, khan2022hardware, khan2022fusion, khan2011coherence, khan2010fingerprint}. However, this method assumes a constant local ridge frequency to save computational complexity, which results in imperfect filtering in some places. An efficient enhancement technique based on the Gabor filter, whose form is adjustable by four parameters, was presented by Hong et al. \cite{hong1998fingerprint, yang2003modified}. This method's advantage is that the filter's orientation and frequency are adaptively determined by the orientation and frequency of local ridges. However, in areas where fingerprint valley and ridge patterns diverge from a pure sinusoidal pattern, the filtering performance is sub-par. Yang et al.\cite{yang2003modified} suggested using positive and negative ridge frequencies based on the local valley width and ridge width, respectively, to solve this problem. Zhu et al. \cite{zhu2004fingerprint} suggested employing a circular filter kernel as an alternative to the squared Gabor filter kernel in \cite{hong1998fingerprint, yang2003modified}, which is useful to prevent artefacts in the filtering process. The approaches listed above are mainly concerned with improving contact fingerprints. A brand new technique has been proposed that focusses on improving contactless fingerprints by Yin et al. \cite{yin2016contactless}.
Fourier domain filtering is a commonly employed approach for fingerprint enhancement, in addition to spatial domain filtering methods. These techniques explicitly define filters in the Fourier domain. According to Sherlock et al. \cite{sherlock1994fingerprint}, improve fingerprint images by applying the fast Fourier transform. This method involves multiplying the Fourier transform of the fingerprint image by n precomputed filters. The output of the filter whose orientation is closest to the local ridge orientation determines the enhancement fingerprint's pixel value. The ridge frequency is constant, which is a disadvantage of this approach.

\begin{figure}
    \centering
    \includegraphics[width=\textwidth]{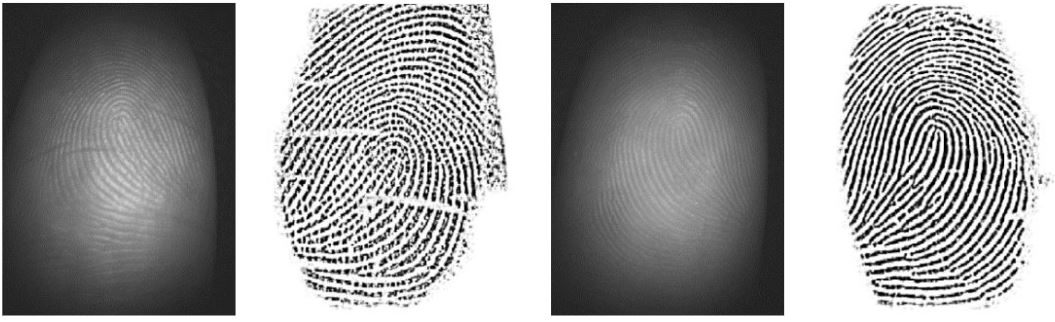}
    \caption{Example images of contactless and contact based fingerprints.}
    \label{fig:contact_contactless}
\end{figure}

\begin{figure*}[h]
    \centering
    \includegraphics[width=0.75\textwidth]{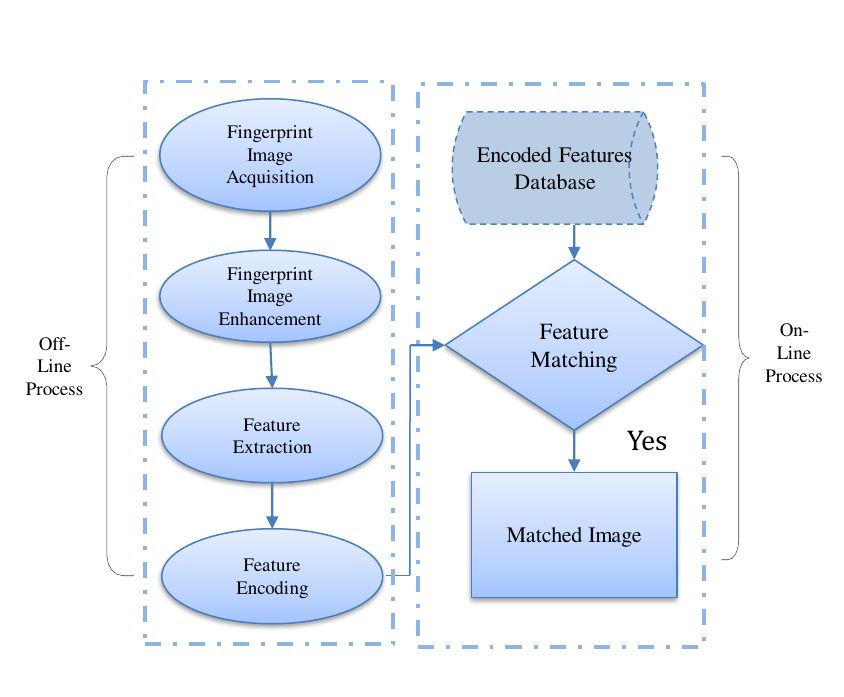}
    \caption{Pipeline of the proposed fingerprint recognition system}
    \label{pipeline}
\end{figure*}

Most of the existing methods focus on enhancing contact fingerprint images and do not account for the unique characteristics of contactless fingerprint images. Unlike conventional fingerprint scanners, this technology operates without physical contact, utilizing advanced scanning methods. Its major benefit became apparent during the pandemic, as it helped reduce the risk of spreading infectious diseases. Compared to contact fingerprint images, contactless fingerprint images exhibit four distinct properties: (1) they have less noise, as illustrated in Figure \ref{fig:contact_contactless}; (2) they have fewer ridge discontinuities, which facilitates the filtering process for image enhancement; (3) the ridge-valley contrast is significantly less distinct, and (4) present an interoperability issue because they do not include the elastic deformation that occurs when pressing a finger against the capture device, complicating the enhancement process. We propose a robust algorithm for contactless fingerprint enhancement and matching to address these differences and leverage the properties of contactless fingerprint images. The experimental results demonstrate the effectiveness of the proposed method in enhancing contactless fingerprint images.

The remainder of this paper is structured as follows. Section \ref{method} details the proposed approach. Section \ref{results} presents the results and discussion. Finally, Section \ref{conclusion} concludes the paper.

\section{Proposed Methodology}
\label{method}

The proposed fingerprint identification algorithm consists of two main phases as shown in Figure \ref{pipeline}. The first phase, which is offline, involves enhancing fingerprint images, extracting minutiae features, and encoding them. This step is performed once for all images in the database to create an encoded candidate database. The second phase, an online process, includes acquiring a new template image from the sensor, enhancing it, extracting minutiae features, and encoding it. The encoded template is then matched against the pre-encoded candidate database from the first phase. In the proposed enhancement algorithm, the initial step involves segmenting the fingerprint image from the background to obtain the region of interest (ROI). To enhance the ridge/valley structure, Gabor-based contextual filtering is applied, which utilises local ridge frequency and orientation information. The fingerprint image is then enhanced using a Gabor filter, followed by binarization and thinning to produce a ridge/valley skeleton. A quality mask is created to extract high-quality minutiae features from the thinned image. The curvature of the local ridge, determined from the orientation of the local ridges, helps to estimate the quality of the regions. This process selects a limited number of high-quality minutiae features, leading to accurate fingerprint identification and reduced computational complexity during feature matching. The output is a quality-based list of minutiae that is encoded for identification purposes.

\subsection{Fingerprint enhancement}
To enhance a fingerprint image, the first step is to obtain the region of interest (ROI), separating the useful fingerprint region from the background. This segmentation is performed using the average magnitude of the gradient, which is higher in the foreground and lower in the background. After extracting the ROI, local ridge orientations and frequencies are calculated, which are necessary for Gabor-based enhancement. The most common method of calculating the orientations of the local ridges involves the calculation of gradients \cite{maltoni2009handbook}. The proposed algorithm uses a method similar to that proposed by Ratha et al. \cite{ratha1995adaptive}, where the ridge directions are computed using the inverse tangent of the gradient in both horizontal and vertical directions. The result of this step is an orientation image denoted by $O_{x,y}$, which contains the local ridge angles at each point $(x,y)$ of the fingerprint image. Once the ridge directions are computed, the next step is to estimate the local ridge frequencies. Estimating ridge frequency can be challenging due to the presence of scars and broken ridges at minutiae locations. Therefore, the proposed enhancement algorithm employs a modified version of the $x-signature$ method \cite{hong1998fingerprint}. In this method, the fingerprint image, sized $(M\times N)$, is divided into $B$ blocks of size $(b\times b)$. Each block is rotated by the angle $O_{m,n}$ to align the ridges vertically, where $m$ and $n$ are the block indices in the vertical and horizontal directions, respectively. Then each block is divided into equal $S$ segments, and the vertical projection of each segment is computed. The frequency of each segment is determined by counting the number of peaks and dividing by the distance between the first and last peaks. The dominant frequency of each block is calculated using the alpha-trimmed mean filtering of the $S$ segment frequencies. This approach provides accurate local ridge frequency estimations even in the presence of high ridge curvature.

\begin{figure*}[h]
    \centering
    \includegraphics[width=\textwidth]{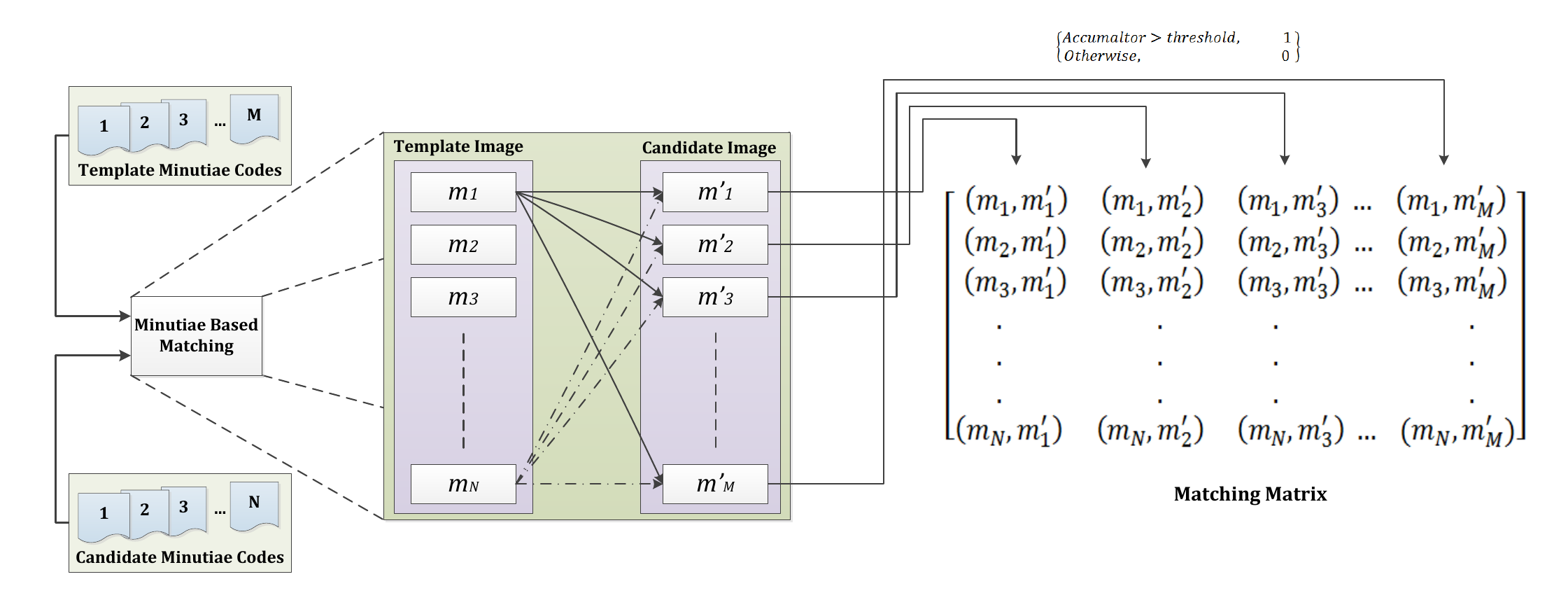}
    \caption{Fingerprint matching based on neighboring Minutia.}
    \label{fig:matching}
\end{figure*}

After calculating the local ridge frequencies and orientations, the fingerprint image can be enhanced using a Gabor filter. The enhancement method, originally proposed by Hong et al. \cite{hong1998fingerprint}, involves convolving each point of the fingerprint image with a Gabor filter tuned to the local ridge orientations and frequencies. The proposed algorithm generates a set of Gabor filters for a range of discretised frequencies and orientations present in the fingerprint image. Each pixel in the fingerprint image is then enhanced using a Gabor filter whose frequency and orientation are closest to the pixel's frequency $P_{f}$ and its orientation $P_{\theta}$, as described by the Eq. \ref{Eq:enh1}

\begin{equation}
    g(x,y:\theta_{p},f_{p})=exp\left[-\frac{1}{2}\left(\frac{x^{2}_{\theta_{p}}}{\sigma^{2}_{x}}+\frac{y^{2}_{\theta_{p}}}{\sigma^{2}_{y}}\right)\right]\cos(2\pi f_{p} x_{\theta_{p}})
    \label{Eq:enh1}
\end{equation}

\begin{equation}
    x_{\theta_{p}}=xco(\theta_{p})+ysin(\theta_{p})
    \label{Eq:enh2}
\end{equation}

\begin{equation}
    y_{\theta_{p}}=-xsin(\theta_{p})+ycos(\theta_{p})
    \label{Eq:enh2}
\end{equation}

where $\sigma_{x}$ and $\sigma_{y}$ are the standard deviations of the Gaussian envelope along the $x$ and $y$ axes, respectively. The resulting enhanced image is then converted into a ridge/valley skeleton to facilitate the extraction of minutiae features. To achieve this skeletonization, the enhanced image undergoes binarization and thinning. Binarization involves converting the second enhanced image into a binary image by clasifying all pixels as either ones or zeroes based on a selected threshold. The binarized image is then thinned to reduce the ridge width to a single pixel, enabling reliable extraction of minutiae features. The most commonly used minutiae features are ridge endings and bifurcations, which can be easily extracted from the thinned fingerprint image. Figure \ref{fig:enhancement} illustrates the results of the enhancement steps and their visual representation on a sample fingerprint image from the FVC2002-DB1A database. Before extracting these minutiae features, a quality mask is created and used in conjuction with the thinned image to extract minutiae from high-quality regions.

\subsection{Feature Extraction}

Fingerprint images encompass various features crucial for fingerprint matching, such as ridges, singular points (core points), and minutiae. The most reliable feature for fingerprint matching is the minutiae point, first observed by Sir Francis Galton based on the discontinuities in local ridge patterns \cite{jain1997line}, hence termed "Galton details". Each minutiae point possesses its location. Ridge endings and ridge bifurcations are primarily used for fingerprint identification due to their stability and accurate detection compared to other minutiae types. Moreover, all other minutiae points can be understood as combinations of ridge endings and bifurcations, obviating the need to detect them separately. These minutiae points are typically extracted from thinned fingerprint images. However, thinning algorithms can introduce noise in the ridge pattern, leading to the detection of spurious minutiae.

To extract candidate minutiae points, a local neighborhood of each pixel is scanned within a $3 \times 3$ window. A ridge pixel is classified on the basis of the number of transitions from 0 to 1 that occur during the scanning of the eight neighbouring pixels of a candidate minutiae in a clockwise direction. A pixel is identified as a ridge bifurcation if there are 3 transitions from 0 to 1, whereas a pixel is identified as a ride ending if only 1 transition from 0 to 1 is detected. Paul et al. \cite{tico2000algorithm} introduced a technique that analyses each candidate's minutiae within a window of size $W\times W$ in the thinned image. After minutiae feature extraction and false minutiae removal steps, a minutiae list is obtained. Each minutiae point is represented by its location and dominant angle.

\begin{figure*}[h]
    \centering
     \resizebox{1.0\textwidth}{!}{%
     \begin{tabular}{@{}c@{\ }c@{\ }c@{\ }c@{\ }c@{\ }c@{\ }c@{\ }c@{}}        
        \includegraphics[width=0.15\textwidth]{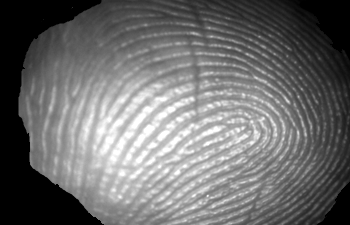} &        \includegraphics[width=0.15\textwidth]{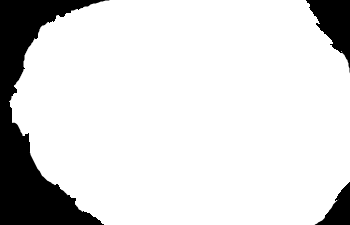} &        \includegraphics[width=0.15\textwidth]{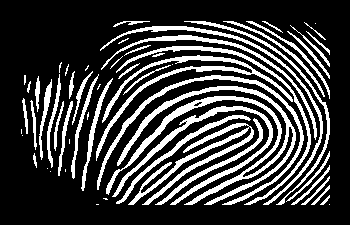} &        \includegraphics[width=0.15\textwidth]{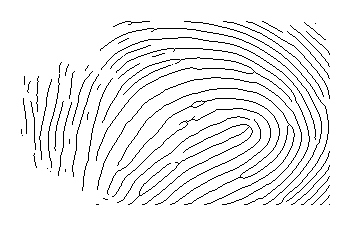} &        \includegraphics[width=0.15\textwidth]{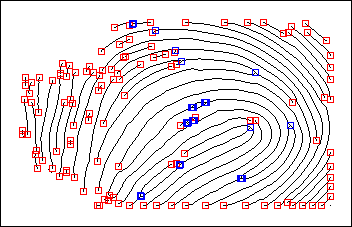} \\

        \includegraphics[width=0.15\textwidth]{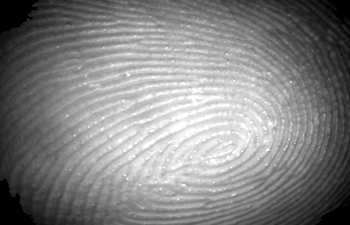} &        \includegraphics[width=0.15\textwidth]{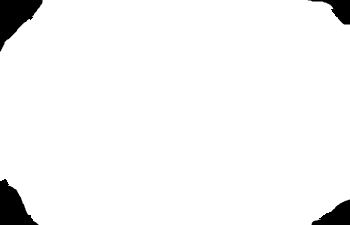} &        \includegraphics[width=0.15\textwidth]{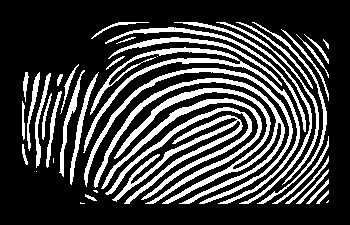} &        \includegraphics[width=0.15\textwidth]{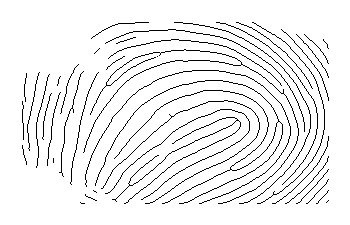} &        \includegraphics[width=0.15\textwidth]{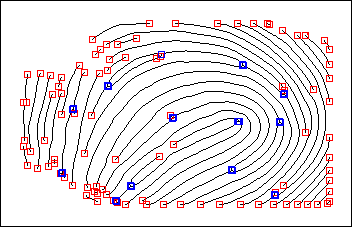} \\
        
        \includegraphics[width=0.15\textwidth]{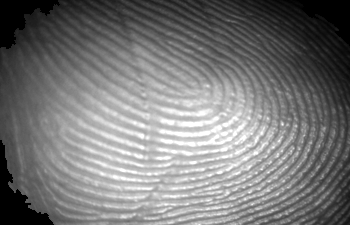} &        \includegraphics[width=0.15\textwidth]{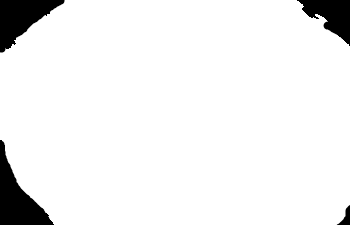} &        \includegraphics[width=0.15\textwidth]{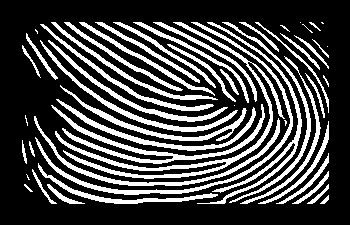} &        \includegraphics[width=0.15\textwidth]{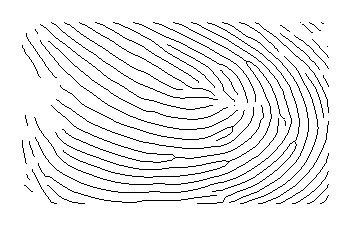} &        \includegraphics[width=0.15\textwidth]{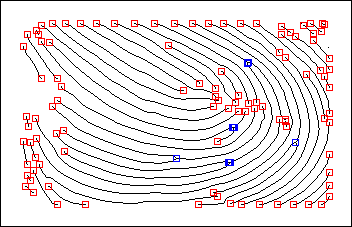} \\

    \end{tabular}
    }
    \caption{Visual performance of the proposed method. From left to right input image, the corresponding mask, binary image, thinned image, and minutiae marked on thinned image are illustrated.}
    \label{fig:minute_detection}
\end{figure*}

\subsection{Feature Encoding}
Once the extraction of features from the fingerprint image is complete, the feature encoding process is initiated. In this stage, each minutiae point is selected as a reference point and encoded based on its neighboring minutiae. The following procedure is typically employed to encode a fingerprint image:

\begin{enumerate}
    \item For each minutiae point minutiae ($M_{i}$) from the minutiae list ($M_{L}$), where i is the index of the reference minutiae point.
    \begin{itemize}
        \item Based on the Euclidean distance between the minutiae ($M_{i}$) and all other minutiae in the minutiae list ($M_{L}$), create a new list ($M^{Nearest}_{i}$) of $n$ nearest neighbouring minutiae points, excluding the reference minutiae ($M_{i}$) itself. This can be expressed a follows;
        \begin{equation}
            M^{Nearest}_{i}(x_{k}, y_{k}, \theta_{k}) = M_{L}(x_{j}, y_{j}, \theta_{j})
        \end{equation}

        Where $k=1:n$ iterates through the indices of the nearest minutiae points, and $j$ represents the index of each nearest minutiae point from the minutiae list ($M_{L}$).

        \item An encoded list of minutiae, denoted as ($m_{i})$, is generated by considering the relative distance and angle $(\rho_{i}, \theta_{i}, \phi_{i})$ features between the focal minutiae ($M_{i}$) and its neighboring minutiae ($M^{Nearest}_{i}$). The features $(\rho_{j}, \theta_{j}, \phi_{j})$ for each minutiae are computed as given in Eq. (\ref{Eq:E1}-\ref{Eq:E3}).
        \begin{equation}
            \rho_{j} = \sqrt{dx^{2}_{j} + dy^{2}_{j}}
            \label{Eq:E1}
        \end{equation}

        \begin{equation}
            \theta_{j}=\tan^{-1}(\frac{dy_{j}}{dx_{j}})
            \label{Eq:E2}
        \end{equation}

        \begin{equation}
            \phi_{j}=M_{L}(3\times (i-1)+2) - M^{Nearest}-{i}(3\times (j-1)+2)
            \label{Eq:E3}
        \end{equation}

        where,
        \begin{equation}
            dx_{j}=M_{L}(3\times(i-1))-M^{Nearest}_{i}(3\times(j-1))
            \label{Eq:E4}
        \end{equation}
        \begin{equation}
            dy_{j}=M_{L}(3\times(i-1)+1)-M^{Nearest}_{i}(3\times(j-1)+1)
            \label{Eq:E5}
        \end{equation}

        The encoding of the minutiae ($m_{i}$) involves a combination of features $(\rho_{j},\theta_{j},\phi_{j})$ derived from the minutiae $n$ nearest-neighbours, excluding the reference minutiae itself, as described in the above step.
    \end{itemize}
    \item The final list of encoded minutiae ($M^{Encoded}_{L}$) for a given fingerprint comprises a combination of all individual minutiae encodings $m_{i}$, represented in Eq. \ref{Eq:E6}.
    \begin{equation}
        M^{Encoded}_{L}={m_{1}, m_{2}, m_{3}, ..., m_{N}}
        \label{Eq:E6}
    \end{equation}
    where $N$ is the number of minutiae points in a given fingerprint image.

    \item The encoded minutiae list $M^{Encoded}_{L}$ is stored in the database as a candidate finger-code($C$).
\end{enumerate}

\begin{table*}[h]
  \centering
  \caption{EER comparisons of different $n$-nearest neighbors for different matched neighbors PolyU contactless fingerprint database \cite{polyU_dataset}.}
   \adjustbox {max width=\textwidth}{%
    \begin{tabular}{ccccccccccc}
    \toprule
    \multicolumn{1}{c}{\multirow{3}[4]{*}{\textbf{Matched Neighbors}}} & \multicolumn{10}{c}{\textbf{n -  Nearest Neighbors}} \\
\cmidrule{2-11}          & \multirow{2}[2]{*}{\textbf{1}} & \multirow{2}[2]{*}{\textbf{2}} & \multirow{2}[2]{*}{\textbf{3}} & \multirow{2}[2]{*}{\textbf{4}} & \multirow{2}[2]{*}{\textbf{5}} & \multirow{2}[2]{*}{\textbf{6}} & \multirow{2}[2]{*}{\textbf{7}} & \multirow{2}[2]{*}{\textbf{8}} & \multirow{2}[2]{*}{\textbf{9}} & \multirow{2}[2]{*}{\textbf{10}} \\
          &       &       &       &       &       &       &       &       &       &  \\
    \midrule
    \textbf{1} & 18.22 & 18.27 & 19.29 & 22.21 & 23.98 & 27.20 & 29.81 & 32.25 & 33.51 & 36.35 \\
    \textbf{2} & -     & 9.22  & 4.85  & 3.77  & 3.75  & 4.10  & 4.66  & 5.34  & 6.00  & 7.46 \\
    \textbf{3} & -     & -     & 6.73  & 3.59  & 3.25  & 3.44  & 3.34  & 3.04  & 3.16  & 3.33 \\
    \textbf{4} & -     & -     & -     & 11.19 & 4.09  & 3.36  & 3.30  & 3.68  & 2.98  & 3.57 \\
    \textbf{5} & -     & -     & -     & -     & 17.37 & 7.11  & 3.14  & 3.01  & \textcolor{red}{\textbf{2.84}} & 2.98 \\
    \textbf{6} & -     & -     & -     & -     & -     & 22.17 & 10.72 & 5.40  & 3.27  & 3.26 \\
    \textbf{7} & -     & -     & -     & -     & -     & -     & 76.64 & 14.39 & 8.19  & 5.14 \\
    \textbf{8} & -     & -     & -     & -     & -     & -     & -     & 80.82 & 18.64 & 10.55 \\
    \textbf{9} & -     & -     & -     & -     & -     & -     & -     & -     & 83.77 & 22.17 \\
    \textbf{10} & -     & -     & -     & -     & -     & -     & -     & -     & -     & 85.89 \\
    \bottomrule
    \end{tabular}%
    }
  \label{tab:matching_neighbors}%
\end{table*}%

\subsection{Feature Matching}
To match a candidate fingerprint image with template fingerprint images stored in the database, the candidate fingerprint image follows the same steps as the template fingerprint image in preprocessing, enhancement, extraction of minutiae, and encoding. An exhaustive search algorithm is employed for the matching, where the biometric code of each microlevel point of the candidate fingerprint image is matched with all the biometric codes of the template fingerprint image to find the best match between them, as shown in (Fig. \ref{fig:matching}).

Every minutiae code of the candidate is matched with all minutiae codes of the template fingerprint. Due to any spurious or missing neighboring minutiae, the order of occurrence of neighboring minutiae code is not certain. Matching every neighboring minutiae code of candidate minutiae with every neighboring minutiae code of template minutiae is required to find the correct correspondence. To match a candidate minutiae code that consists of $n$ neighbours with one template minutiae code, $n^2$ neighbour matching is required. The accuracy of the matching algorithm is directly proportional to the number of neighboring minutiae for encoding. If the number of neighbors for encoding increases the accuracy of the matching algorithm, the number of computations involved in matching and the size of the biometric code also increases. The overall computational complexity of the minutiae neighbour matching algorithms is $M\times N\times n^{2}$ where $n$ denotes the number of neighbours and M and N are the number of minutiae points in the candidate and template fingerprint images. A similarity score is calculated based on the matching of pairs of minutiae between two fingerprint images.

\section{Results and Discussions}
\label{results}
In this section, we briefly present the results of each stage of the proposed method. First, we have discussed the results of preprocessing stage where the image is enhanced followed by the minutiae extraction and minutiae matching stage results. Finally, we have compared the matching results with the other recent methods.

Figure \ref{fig:minute_detection} presents the results of the detection of minutiae on an enhanced contactless fingerprint image from the PolyU contactless fingerprint database \cite{polyU_dataset}. The enhanced image undergoes binarization followed by a thinning process before extracting the minutiae points, as shown in Figure \ref{fig:minute_detection}.

PolyU contactless fingerprint database \cite{polyU_dataset} is used for experiments. FVC \cite{maio2002fvc2002} testing protocol is used for performance evaluation, where an equal error rate (ERR) performance indicator is used. The total number of fingerprint images (subjects) is 336, and each subject has 6 samples. For EER calculation, the experiments are carried out in two stages, where the false non-matching rate (FNMR) and false matching rate (FMR) are computed. In the first stage, the false non-matching rate (FNMR), which is known as genuine matching, is computed by matching each sample of a subject (fingerprint) with the remaining samples of the same subject. 
In the second stage, the false matching rate (FMR), which is known as imposter matching, is computed by matching the first sample of each subject (fingerprint) with the first sample of the remaining subjects.

The total number of genuine matching ($E_{genuine}$) and imposter matching ($E_{imposter}$) experiments is given in Eq. (\ref{Eq:I1}-\ref{Eq:I2}):

\begin{equation}
    E_{genuine}=336 \times \left ( \frac{6\times 5}{2} \right ) = 10,080
    \label{Eq:I1}
\end{equation}

\begin{equation}
   E_{imposter}=\left ( \frac{336\times 335}{2} \right )=56,280
    \label{Eq:I2}
\end{equation}

EER is a measure of the performance of the system and is given by FMR or FNMR, at a point where both FMR and FNMR are equal \cite{maltoni2009handbook}. Table \ref{tab:matching_neighbors} shows the different EERs achieved based on the different numbers of neighbours matched and the number of neighbours closest $n$ minutiae that are used in the encoding of a given minutia code. Table \ref{tab:matching_neighbors}  shows a complete empirical analysis of EER. It can be seen that as the number of neighbors $n$ involved in encoding increases, the accuracy of the matching algorithm also increases. However, increasing $n$ increases the number of computations involved in matching, as well as the size of the finger code. It can be seen that a minimum EER of 2.84\% is achieved for $k=9$ closest neighbours and a matching neighbour threshold $5$.

\begin{table}[htbp]
  \centering
  \caption{EER and time comparison of the proposed method with other matching algorithms on PolyU contactless fingerprint database \cite{polyU_dataset}.}
  \adjustbox {max width=\textwidth}{%
    \begin{tabular}{lccc}
    \toprule
    \textbf{Method} & \textbf{\# of  Matches} & \textbf{EER in (\%)} & \textbf{(Matches/Sec)} \\
    \midrule
    Chikkerur et al. \cite{chikkerur2005k} & 66,360 & 3.46  & 0.98 \\
    NIST Bozorth3 \cite{watsonuser} & 66,360 & 4.74  & 188  \\
    \midrule
    \textbf{Proposed} & 66,360& \textbf{2.84} & \textbf{206} \\
    \bottomrule
    \end{tabular}%
    }
  \label{tab:comparison}%
\end{table}%

Table \ref{tab:comparison} presents the EER comparison of the proposed method with other matching algorithms (K-plet and coupled BFS proposed by Chikkerur et al. \cite{chikkerur2005k} and NIST Bozorth3 \cite{watsonuser}) respectively. It is evident from Table \ref{tab:comparison} that the proposed method has outperformed all other methods in terms of EER by achieving 0.6\% and  1.9\% times better EER than the K-plet and coupled BFS proposed by Chikkerur et al. \cite{chikkerur2005k} and NIST Bozorth3 \cite{watsonuser}, respectively.

\section{Conclusions}
\label{conclusion}

In this study, we introduced a solution for contactless fingerprint identification. The primary objective of the contactless fingerprint enhancement was to improve the clarity of pertinent details while mitigating noise artefacts within the fingerprint images, thus improving the accuracy of identification. Once the fingerprint image is enhanced, the proposed method uses minutiae extraction, followed by the minutiae encoding stage. Subsequently, a minutiae-based matching algorithm was employed. Finally, the experimental findings validate the significant superiority of our proposed approach to fingerprint enhancement and matching. The proposed minutiae matching algorithm achieved a minimum EER of 2. 84\% on the PolyU contactless fingerprint database. Future research directions may include GAN-based fingerprint enhancement. Additionally, investigating the integration of GAN with existing forensic tools and workflows could facilitate seamless adoption and integration into real-world applications.

\bibliographystyle{IEEEtran}
\bibliography{References}

\begin{thebibliography}{10}
\providecommand{\url}[1]{#1}
\csname url@samestyle\endcsname
\providecommand{\newblock}{\relax}
\providecommand{\bibinfo}[2]{#2}
\providecommand{\BIBentrySTDinterwordspacing}{\spaceskip=0pt\relax}
\providecommand{\BIBentryALTinterwordstretchfactor}{4}
\providecommand{\BIBentryALTinterwordspacing}{\spaceskip=\fontdimen2\font plus
\BIBentryALTinterwordstretchfactor\fontdimen3\font minus
  \fontdimen4\font\relax}
\providecommand{\BIBforeignlanguage}[2]{{%
\expandafter\ifx\csname l@#1\endcsname\relax
\typeout{** WARNING: IEEEtran.bst: No hyphenation pattern has been}%
\typeout{** loaded for the language `#1'. Using the pattern for}%
\typeout{** the default language instead.}%
\else
\language=\csname l@#1\endcsname
\fi
#2}}
\providecommand{\BIBdecl}{\relax}
\BIBdecl

\bibitem{michael2012contactless}
G.~Michael, T.~Connie, and A.~Teoh, ``A contactless biometric system using
  multiple hand features,'' \emph{Journal of Visual Communication and Image
  Representation}, vol.~23, no.~7, pp. 1068--1084, 2012.

\bibitem{michael2010innovative}
G.~Michael, T.~Connie, and A.~Teoh Beng~Jin, ``An innovative contactless palm
  print and knuckle print recognition system,'' \emph{Pattern Recognition
  Letters}, vol.~31, no.~12, pp. 1708--1719, 2010.

\bibitem{oh2017gabor}
B.~Oh, K.~Oh, A.~Teoh, Z.~Lin, and K.~Toh, ``A gabor-based network for
  heterogeneous face recognition,'' \emph{Neurocomputing}, vol. 261, pp.
  253--265, 2017.

\bibitem{zhou2014benchmark}
W.~Zhou, J.~Hu, I.~Petersen, S.~Wang, and M.~Bennamoun, ``A benchmark 3d
  fingerprint database,'' in \emph{2014 11th International Conference on Fuzzy
  Systems and Knowledge Discovery (FSKD)}.\hskip 1em plus 0.5em minus
  0.4em\relax IEEE, 2014, pp. 935--940.

\bibitem{khan2013fingerprint}
M.~A. Khan and T.~M. Khan, ``Fingerprint image enhancement using data driven
  directional filter bank,'' \emph{Optik}, vol. 124, no.~23, pp. 6063--6068,
  2013.

\bibitem{khan2014fingerprint}
T.~M. Khan, M.~A. Khan, and Y.~Kong, ``Fingerprint image enhancement using
  multi-scale ddfb based diffusion filters and modified hong filters,''
  \emph{Optik}, vol. 125, no.~16, pp. 4206--4214, 2014.

\bibitem{khan2016spatial}
M.~A. Khan, T.~M. Khan, D.~G. Bailey, and Y.~Kong, ``A spatial domain scar
  removal strategy for fingerprint image enhancement,'' \emph{Pattern
  Recognition}, vol.~60, pp. 258--274, 2016.

\bibitem{khan2016stopping}
T.~M. Khan, M.~A. U.~Khan, Y.~Kong, and O.~Kittaneh, ``Stopping criterion for
  linear anisotropic image diffusion: a fingerprint image enhancement case,''
  \emph{EURASIP Journal on Image and Video Processing}, vol. 2016, pp. 1--20,
  2016.

\bibitem{khan2017efficient}
T.~M. Khan, D.~G. Bailey, M.~A. Khan, and Y.~Kong, ``Efficient hardware
  implementation for fingerprint image enhancement using anisotropic gaussian
  filter,'' \emph{IEEE Transactions on Image processing}, vol.~26, no.~5, pp.
  2116--2126, 2017.

\bibitem{khan2018coupling}
M.~A.~U. Khan, T.~M. Khan, and D.~G. Bailey, ``Coupling orientation diffusion
  with coherence-enhancing diffusion: a fingerprint case,'' \emph{Signal, Image
  and Video Processing}, vol.~12, no.~3, pp. 513--521, 2018.

\bibitem{chikkerur2007fingerprint}
S.~Chikkerur, A.~Cartwright, and V.~Govindaraju, ``Fingerprint enhancement
  using stft analysis,'' \emph{Pattern Recognition}, vol.~40, no.~1, pp.
  198--211, 2007.

\bibitem{sherlock1994fingerprint}
B.~Sherlock, D.~Monro, and K.~Millard, ``Fingerprint enhancement by directional
  fourier filtering,'' \emph{IEE Proceedings - Vision, Image and Signal
  Processing}, vol. 141, no.~2, pp. 87--94, 1994.

\bibitem{watson1994comparison}
C.~Watson, G.~Candela, and P.~Grother, ``Comparison of fft fingerprint
  filtering methods for neural network classification,'' National Institute of
  Standards and Technology (NIST), Technical Report NISTIR 5493, 1994.

\bibitem{nickerson1989approach}
J.~Nickerson and L.~O’Gorman, ``An approach to fingerprint filter design,''
  \emph{Pattern Recognition}, vol.~22, no.~1, pp. 29--38, 1989.

\bibitem{sabir2020reducing}
M.~Sabir, T.~Khan, M.~Arshad, and S.~Munawar, ``Reducing computational
  complexity in fingerprint matching,'' \emph{Turkish Journal of Electrical
  Engineering and Computer Sciences}, vol.~28, no.~5, pp. 2538--2551, 2020.

\bibitem{khan2022hardware}
T.~M. Khan, ``Hardware implementation of multimodal biometric using fingerprint
  and iris,'' \emph{arXiv preprint arXiv:2201.05996}, 2022.

\bibitem{khan2022fusion}
------, ``Fusion of fingerprint and iris recognition for embedded multimodal
  biometric system,'' Ph.D. dissertation, Macquarie University, 2022.

\bibitem{khan2011coherence}
M.~A. Khan, T.~M. Khan, and S.~A. Khan, ``Coherence enhancement diffusion using
  multi-scale dfb,'' in \emph{2011 7th International Conference on Emerging
  Technologies}.\hskip 1em plus 0.5em minus 0.4em\relax IEEE, 2011, pp. 1--6.

\bibitem{khan2010fingerprint}
M.~A. Khan, A.~Khan, T.~Mahmood, M.~Abbas, and N.~Muhammad, ``Fingerprint image
  enhancement using principal component analysis (pca) filters,'' in \emph{2010
  International Conference on Information and Emerging Technologies}.\hskip 1em
  plus 0.5em minus 0.4em\relax IEEE, 2010, pp. 1--6.

\bibitem{hong1998fingerprint}
L.~Hong, Y.~Wan, and A.~Jain, ``Fingerprint image enhancement: algorithm and
  performance evaluation,'' \emph{IEEE Transactions on Pattern Analysis and
  Machine Intelligence}, vol.~20, no.~8, pp. 777--789, 1998.

\bibitem{yang2003modified}
J.~Yang, L.~Liu, T.~Jiang, and Y.~Fan, ``A modified gabor filter design method
  for fingerprint image enhancement,'' \emph{Pattern Recognition Letters},
  vol.~24, no.~12, pp. 1805--1817, 2003.

\bibitem{zhu2004fingerprint}
E.~Zhu, J.~Yin, and G.~Zhang, ``Fingerprint enhancement using circular gabor
  filter,'' in \emph{International Conference on Image Analysis and
  Recognition}, ser. Lecture Notes in Computer Science, A.~Campilho and
  M.~Kamel, Eds., vol. 3212.\hskip 1em plus 0.5em minus 0.4em\relax Heidelberg:
  Springer, 2004, pp. 750--758.

\bibitem{yin2016contactless}
X.~Yin, J.~Hu, and J.~Xu, ``Contactless fingerprint enhancement via intrinsic
  image decomposition and guided image filtering,'' in \emph{2016 IEEE 11th
  Conference on Industrial Electronics and Applications (ICIEA)}.\hskip 1em
  plus 0.5em minus 0.4em\relax IEEE, 2016, pp. 144--149.

\bibitem{maltoni2009handbook}
D.~Maltoni, D.~Maio, A.~K. Jain, S.~Prabhakar \emph{et~al.}, \emph{Handbook of
  fingerprint recognition}.\hskip 1em plus 0.5em minus 0.4em\relax Springer,
  2009, vol.~2.

\bibitem{ratha1995adaptive}
N.~K. Ratha, S.~Chen, and A.~K. Jain, ``Adaptive flow orientation-based feature
  extraction in fingerprint images,'' \emph{Pattern recognition}, vol.~28,
  no.~11, pp. 1657--1672, 1995.

\bibitem{polyU_dataset}
C.~Lin and A.~Kumar, ``Matching contactless and contact-based conventional
  fingerprint images for biometrics identification,'' \emph{IEEE Transactions
  on Image Processing}, vol.~27, no.~4, pp. 2008--2021, 2018.

\bibitem{maio2002fvc2002}
D.~Maio, D.~Maltoni, R.~Cappelli, J.~L. Wayman, and A.~K. Jain, ``Fvc2002:
  Second fingerprint verification competition,'' in \emph{2002 International
  conference on pattern recognition}, vol.~3.\hskip 1em plus 0.5em minus
  0.4em\relax IEEE, 2002, pp. 811--814.

\bibitem{chikkerur2005k}
S.~Chikkerur, A.~N. Cartwright, and V.~Govindaraju, ``K-plet and coupled bfs: a
  graph based fingerprint representation and matching algorithm,'' in
  \emph{Advances in Biometrics: International Conference, ICB 2006, Hong Kong,
  China, January 5-7, 2006. Proceedings}.\hskip 1em plus 0.5em minus
  0.4em\relax Springer, 2005, pp. 309--315.

\bibitem{watsonuser}
C.~I. Watson, M.~D. Garris, E.~Tabassi, C.~L. Wilson, R.~M. McCabe, S.~Janet,
  and K.~Ko, ``User's guide to,'' \emph{NIST Biometric Image Software}.

\end{thebibliography}

\end{document}